# Using Large Language Models to Generate Engaging Captions for Data Visualizations

Ashley Liew*  
Rochester Institute of Technology

Klaus Mueller†  
Stony Brook University

## ABSTRACT

Creating compelling captions for data visualizations has been a long-standing challenge. Visualization researchers are typically untrained in journalistic reporting and hence the captions that are placed below data visualizations tend to be not overly engaging and rather just stick to basic observations about the data. In this work we explore the opportunities offered by the newly emerging crop of large language models (LLM) which use sophisticated deep learning technology to produce human-like prose. We ask, can these powerful software devices be purposed to produce engaging captions for generic data visualizations like a scatterplot. It turns out that the key challenge lies in designing the most effective prompt for the LLM, a task called *prompt engineering*. We report on first experiments using the popular LLM GPT-3 and deliver some promising results.

**Index Terms:** Human-centered computing—Visualization—Prompt engineering; GPT-3—Natural language model—Captioning

## 1 INTRODUCTION

Captions of figures form an important part in the reporting of scientific data. They aid readers in understanding and correctly interpreting the figure. Many guides on scientific writing advise that a graph or image in a report is incomplete without a proper caption. It is also often mentioned that a caption should be able to stand on its own. It turns out that text plays an important part in a data visualization; a study by Borkin et al. [1] indicated that the textual elements of a visualization, including 'title' and 'paragraph' had some of the highest recall rates. Other work in visualization (such as [2] and many papers that followed) revealed that making visualizations more engaging can also make them more memorable. Finally, when it comes to the text itself it should be obvious that an engaging text will be more attractive to a reader than a simple one that just blandly states some statistics extracted from the visualization.

But writing an engaging caption is difficult. And so we ask, can we build a creative writing assistant to help in composing a caption that reports on the facts and statistics, verifiable from the visualization, and at the same time also adds interesting journalistic content, even commentary, that enlivens the caption and makes it a complementary partner to the visual depiction of the data. This has motivated us to research the use of large language models such as GPT-3 (short for Generative Pre-trained Transformer 3), which have been shown to be capable of creative writing – producing essays, dialog, and even poems that rival the capabilities of humans.

Our paper is structured as follows. Section 2 presents related work. Section 3 presents our general approach, along with technical details on the data visualizations we used and the prompt generation, which forms an important element of our work. Section 4 lists the observations we found along with some of the outputs we obtained from GPT-3. Section 5 describes a user study we conducted and Section 6 discusses our thoughts on the results. Section 7 ends with an outlook on future work.

## 2 RELATED WORK

Visualization has a strong connection to storytelling but thus far the primary focus has been on storytelling with visuals, and visuals alone, possibly annotated [3]. The frequently cited paper by Segel and Heer [4] proposes seven genres of narrative visualization, such as flow chart, comic strip, slide show, and others that all focus on the ordering and organization of visuals to tell the story. Many papers have followed this seminal work and a good overview is presented in the paper by Tong et al. [7].

On the other hand, natural language interfaces have also become popular recently in the field of visualization. The aim here is to empower a user to describe a certain visualization goal verbally and the system would produce a visualization fitting this goal. They free the user from manipulating possibly complex interface elements and instead prompt the system to deliver the right visualization for the aim stated verbally. The paper by Shen et al. [5] provides a good survey on these types of systems.

While there has not been much work on automated caption generation in the field of visualization – most often simple templates are used – the task of describing images with syntactically and semantically meaningful sentences has seen a rich body of research (see the survey by Stefanini et al. [6]). Related are the products of the company Narrative Science; they analyze structured data and then automatically generate textual narratives which can be accompanied by a visualization of these data.

## 3 APPROACH

GPT-3 is part of a new generation of task-agnostic large language models that are capable of numerous natural language tasks including translation, summarization, and text generation. To interface with these models, users provide an input prompt describing the task at hand for which it will then generate an appropriate output. The design of the prompt, coupled with the model hyperparameter settings (how hyperparameters fit into our study is mentioned later in this section), is crucial for producing relevant and quality outputs and is culminated into the field of prompt engineering. Several techniques, such as few-shot learning or chain reasoning, have garnered attention for their usefulness in generating such outputs, however they also require significant amount of human effort in designing the input prompt. We focus our type of prompt engineering on the amount of human interaction with GPT-3, i.e. how involved the user of the creative writing agent must be to produce their desired output. This simulates the actual process of caption generation: analysts, researchers, or amateur data journalists who are more active will also be able to create more engaging and compelling captions than someone who is less.

Thus, our study can be broken down into two parts: developing a baseline approach to generate captions for data visualization and studying how to extend this baseline with more human effort prompt engineering to increase its engagement. We start out with conducting simple analysis, either linear regression or clustering, on a data set to obtain certain measurements; Table 1 shows the type of metadata collected for each analysis method. Using a three-level tier-based approach of prompt engineering, we populate a template with the collected measurements (first part) and for each subsequent tier, we extend the template with additional prompt engineering (second part). At the end, we have several data visualizations and its three captions produced from varying amounts of prompt engineering. To empirically test how engaging the captions are and compare each tier, we conduct a user study.

---

*e-mail: acl9213@rit.edu  
†e-mail: mueller@cs.stonybrook.edu

| Both | |
|---|---|
| Title, Axes labels, Value ranges, Other columns in the data set | |
| **Linear Regression** | **Clustering** |
| Regression coefficients, Outliers by studentized residual distances, Pearson's correlation | Number of clusters, Cluster sizes, Description of clusters |

Table 1: Metadata included for each type of analysis

### 3.1 Visualization Generation

The data visualizations used in our experiment were manually produced. Without loss of generality, we limited the scope of our investigations to two-dimensional scatter plot visualizations. Datasets were collected as CSV files from the data repository site Kaggle and ensured to have at least two numeric variables that would be used as the x and y-axes. After minor data processing for proper formatting, each CSV file was run through an analysis script and returned as a two-dimensional scatter plot visualization along with their analysis results.

The analysis script used the Scikit-Learn LinearRegression and DBSCAN models for regression and cluster analysis, respectively. Prior to analysis, each dataset was paired with one of the analysis method based on what we felt was appropriate for its scatter plot. Outliers for linear regression were located using studentized residuals and with any distance greater than three used as the rule of thumb for a possible outlier. As residuals serve as only a possibility of being an outlier, further manual processing was done to remove false positives from the final results. Other measurements recorded for linear regression were its linear equation coefficients, as well as its Pearson's correlation coefficient. Scaling and hyperparameter-tuning was done for DBSCAN datasets to ensure reasonably-sized clusters and sensible outliers. Other measurements recorded for clustering were number of clusters and ranking of cluster sizes. The scatter plot visualizations were generated with Matplotlib's pyplot module and squared based on the 45 Banking Rule for two-dimensional visualizations. Examples of generated visualizations can be viewed in Figure 1. It is important to note that the analysis was not optimized, rather it was done as an means to an end in gathering data visualizations and consistent measurements for our experiment.

### 3.2 Prompt Design and Generation

A broad overview of GPT-3's internal mechanism poses a few constraints to our development of a prompt for caption generation. GPT-3 works in a "next-token" generative process. After an input prompt is submitted, GPT-3 tokenizes it and calculates probabilities of candidate tokens that would continue off the last token. GPT-3 will then generate up to a user-set number of tokens, including the input prompt, for the output i.e. input prompt is counted as part of the token limit. The trade-off for a larger number of tokens is that more money and time is needed. Consequently, a prompt needed to include an optimal amount of detail of the task and avoid overextending the token count. Preliminary experimentation also shows that GPT-3 does not fare well with numeric tasks, such as math operations and aggregation of given data points. Putting these limitations in the context of data visualization captioning, the prompt must contain post-analysis measurements (not requiring GPT-3 to do any numeric analysis itself) and be in natural language.

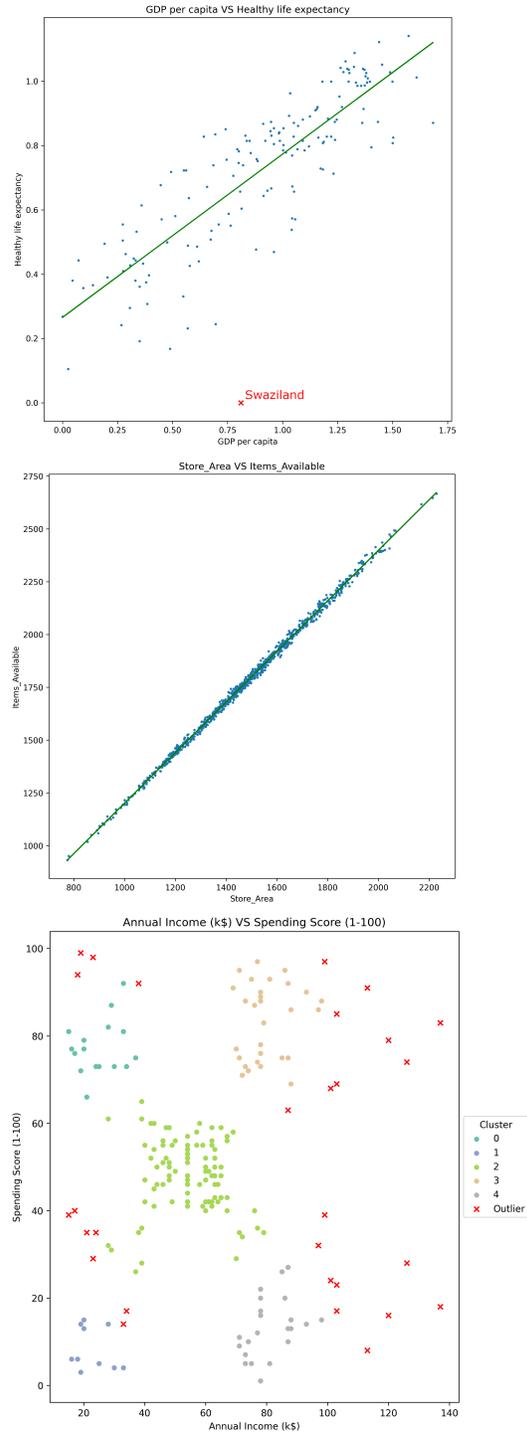

Figure 1: Linear Regression (top 2) and Cluster (last) visualizations

From this, we develop our template baseline (Tier 1). Each analysis method has its own template based on the different metadata (Table 1) they collect. This baseline requires none to minimal human effort; most of the information is already included in the analysis or requires briefly describing what is shown in the visualization (e.g. cluster description is needed). The next tier (Tier 2) of human effort prompt engineering is appending an instructional sentence to the template that tells GPT-3 what specific talking points for the captions to go into (e.g. outliers, correlations). Human effort is needed as it does require more thought into what is the main point of the visualization and its caption, but remains low. The last tier (Tier 3) requires the most amount of human effort. Extending the template and instruction, users can actively adjust the generated captions via question-answering. This prompt engineering approach turns into a near-chatbot interface, and requires that the user is knowledgeable about the data set domain in order to produce their desired output. In addition to questions, users can direct GPT-3 to modify grammatical and structural parts of the captions.

| Tier # | Tier Name | Includes (in addition to previous tier) | Level of Human Effort |
|---|---|---|---|
| 1 | Template Only | Metadata: Labels, Ranges, Analysis measurements and descriptions (for clusters) | None, Close to None |
| 2 | Template + Instruction | Sentence describing specific talking points | Low |
| 3 | Template + Instruction + QA | User-stated modifications, questions, and clarifications | High |

Table 2: Tiers of Human Effort Prompt Engineering

The following is the prompt of a filled-in template, colored in yellow; the filled-in terms specific for the visualization at hand are bold-faced. The relatively short subsequent text in italics is what GPT-3 returned given this prompt. Next are the subsequent prompt extensions, colored orange for tier 2 and colored red for tier 3; the respective outputs produced by GPT-3 are the non-highlighted italicized text. Note that the later tiers build on the former. A summary of all three tiers is shown in Table 2.

(Corresponding visualization is the first plot in Figure 1)

[Prompt] Generate an engaging caption for a scatter plot titled **GDP per capita VS Healthy life expectancy** with the x-axis labeled as **GDP per capita** and the y-axis labeled as **Healthy life expectancy**. Other columns from the data set include **Social support, Perceptions of corruption, Generosity, Overall rank, Score, Country or region, and Freedom to make life choices**. The range of **GDP per capita** is **0.0 to 1.684**. The range of **Healthy life expectancy** is **0.0 to 1.141**. The linear regression intercept is **0.27** and the slope is **0.51**. The correlation coefficient is **0.84**. Outliers found are **Swaziland** which had a **lower healthy life expectancy** than would be expected of its **GDP per capita**.

[GPT-3, new caption] *The higher the GDP per capita, the higher the healthy life expectancy!*

[Added prompt] The caption should include information explaining causes of the large positive correlation and why there is an outlier in detail.

[GPT-3, new caption] *There is a strong positive correlation between GDP per capita and Healthy life expectancy. A country's GDP per capita is indicative of the average income of its citizens and the overall wealth of the country. A higher GDP per capita generally means that citizens have more disposable income, which can be used to purchase goods and services that improve their health and wellbeing. The outlier in this data is Swaziland, which has a lower healthy life expectancy than would be expected of its GDP per capita. This is likely due to the high prevalence of HIV/AIDS in the country, which has a significant impact on the health of the population.*

[Added prompt] Are there any other reasons why Swaziland has a lower healthy life expectancy?

[GPT3, new caption] *There is a strong positive correlation between GDP per capita and Healthy life expectancy. A country's GDP per capita is indicative of the average income of its citizens and the overall wealth of the country. A higher GDP per capita generally means that citizens have more disposable income, which can be used to purchase goods and services that improve their health and wellbeing. The outlier in this data is Swaziland, which has a lower healthy life expectancy than would be expected of its GDP per capita. This is likely due to the high prevalence of HIV/AIDS in the country, as well as other factors such as poor access to healthcare and sanitation.*

[Added prompt] Why does Swaziland have poor sanitation?

[GPT-3, new caption] *There is a strong positive correlation between GDP per capita and Healthy life expectancy. A country's GDP per capita is indicative of the average income of its citizens and the overall wealth of the country. A higher GDP per capita generally means that citizens have more disposable income, which can be used to purchase goods and services that improve their health and wellbeing. The outlier in this data is Swaziland, which has a lower healthy life expectancy than would be expected of its GDP per capita. This is likely due to the high prevalence of HIV/AIDS in the country, as well as other factors such as poor access to healthcare, sanitation, and nutrition.*

[Added prompt] What is the reason for Swaziland's poor nutrition?

[GPT-3, new caption] *There is a strong positive correlation between GDP per capita and Healthy life expectancy. A country's GDP per capita is indicative of the average income of its citizens and the overall wealth of the country. A higher GDP per capita generally means that citizens have more disposable income, which can be used to purchase goods and services that improve their health and wellbeing. The outlier in this data is Swaziland, which has a lower healthy life expectancy than would be expected of its GDP per capita. This is likely due to the high prevalence of HIV/AIDS in the country, as well as other factors such as poor access to healthcare, sanitation, and nutrition. Swaziland's poor nutrition is likely due to a combination of factors, including poverty, food insecurity, and a lack of access to nutritious foods.*

During experimentation, GPT-3's hyperparameters (temperature, frequency penalty, presence penalty) were set to 0 to isolate as the only factor causing variability in the captions. These hyperparameters enable GPT-3 to take more risks when choosing candidate tokens or dynamically modify their probabilities in token generative process. In the case of temperature, the effect is that GPT-3 becomes more stochastic, one generation is often different from the next and might vary engagment in its captions for the same hyperparameter value. Contrary, we see that a temperature set to 0 outputs the same captions most of the time. Table 3 shows a series of three generations for temperature values 0 and 1. While there are intermediate values between 0 and 1, we chose to stick to 0 to make GPT-3 completely deterministic.

| Temp-erature | Gene-ration # | Caption |
|---|---|---|
| 0 | 1 | The higher the GDP per capita, the higher the healthy life expectancy! |
| 0 | 2 | same as last generation |
| 0 | 3 | same as last generation |
| 1 | 1 | GDP per capita and healthy life expectancy are positively correlated. Countries with higher GDP per capita tend to have citizens with longer life expectancies. |
| 1 | 2 | Scatter plot of GDP per capita vs. healthy life expectancy with linear regression line and correlation coefficient |
| 1 | 3 | As can be seen in the scatter plot, there is a positive relationship between GDP per capita and Healthy life expectancy. This means that as GDP per capita increases, so does Healthy life expectancy. |

Table 3: Temperature: 0 vs 1 for Tier 1 captions

## 4 OBSERVATIONS

One prominent feature of GPT-3 is its ability to produce convincing but false information and the example below shows such situation. This often led to a contradiction between the visualization and the caption, which in turn could affect its engagement level. But given we don't have a clear definition of what engagement means, we chose to leave this discrepancy out of our experimentation. In fact, this is where we believe Tier 3 would be needed to correct the factual content of the caption. The example below shows how the generated captions does not reflect what the visualization shows.

(Corresponding visualization is the middle plot in Fig. 1)

Generate an engaging caption for a scatter plot titled **Store Area VS Items Available** with the x-axis labeled as **Store Area** and the y-axis labeled as **Items Available**. Other columns from the dataset include **Daily Customer Count, Store Sales, Store ID**. The range of **Store Area** is **775** to **2229**. The range of **Items Available** is **932** to **2667**. The linear regression intercept is **3.97** and the slope is **1.2**. The correlation coefficient is **1.0**.

*The more space a store has, the more items they have available!*

The caption should include information explaining causes of the large positive correlation and why there are no outliers in detail.

*There is a strong positive correlation between store area and items available because stores with more space can stock more items. There are no outliers because all stores are similar in size.*

But doesn't the difference in Store Area mean that the stores are not similar in size?

*There is a strong positive correlation between store area and items available because stores with more space can stock more items. There are no outliers because all stores have similar amounts of space to sell items.*

That doesn't make sense, change the caption so that it shows that there is a difference between the store area.

*There is a strong positive correlation between store area and items available because stores with more space can stock more items. There are no outliers because all stores have similar amounts of space to sell items. However, there is a difference in store size, with the range of store area being 775 to 2229.*

Replace the last two sentences that talks about the difference between the store area.

*There is a strong positive correlation between store area and items available because stores with more space can stock more items. There are no outliers because all stores have similar amounts of space to sell items. This is evident by the linear regression equation, with the intercept being 3.97 and the slope being 1.2. The correlation coefficient is 1.0, which further proves the strong positive correlation between store area and items available.*

A similar observation is GPT-3's low precision in satisfying the user's requests. In the example shown in Section 3.2, the user asks *"Why does Swaziland have poor sanitation?"* to entice GPT-3 to include more information related to Swaziland's sanitation, however GPT-3 only appends the prompt with an seemingly unrelated mention of nutrition. We speculate it is either because of an unusual token in GPT-3's generative process or perhaps further prompt engineering is needed for the user-stated questions themselves. Fortunately, these mishaps are not the only things GPT-3 is limited to. Following is an example where GPT-3 responds well.

(Corresponding visualization is the last plot in Figure 1)

Generate an engaging caption for a scatter plot titled **Annual Income (k$) VS Spending Score (1-100)** with the x-axis labeled as **Annual Income (k$)** and the y-axis labeled as **Spending Score (1-100)**. Other columns from the dataset include **Age, CustomerID, Gender**. The range of **Annual Income (k$)** is **15** to **137**. The range of **Spending Score (1-100)** is **1** to **99**. The number of clusters is **5**. The largest cluster has **92 customers with average income and average spending score. Other clusters include low income and low spending score, low income and high spending score, high income and low spending score, and high income and high spending score.**

*How much money do you need to make to be a big spender?*

The caption should include information explaining causes of the large positive correlation and why there are no outliers in detail.

*The clusters in this scatter plot represent different spending habits of customers based on their annual income. The largest cluster, in the middle of the plot, represents customers with average incomes and average spending scores. The other clusters represent customers with low incomes and low spending scores, low incomes and high spending scores, high incomes and low spending scores, and high incomes and high spending scores, respectively. The cause of the clusters is likely due to the fact that customers with higher incomes tend*

*to spend more money than customers with lower incomes.*

What about the cluster with low income and high spending score, or the cluster with high income and low spending score?

*The clusters in this scatter plot represent different spending habits of customers based on their annual income. The largest cluster, in the middle of the plot, represents customers with average incomes and average spending scores. The other clusters represent customers with low incomes and low spending scores, low incomes and high spending scores, high incomes and low spending scores, and high incomes and high spending scores, respectively. The cause of the clusters is likely due to the fact that customers with higher incomes tend to spend more money than customers with lower incomes. However, there are some outliers in the data, such as the cluster of customers with low incomes and high spending scores, or the cluster of customers with high incomes and low spending scores.*

Include reasons for the outliers.

*The clusters in this scatter plot represent different spending habits of customers based on their annual income. The largest cluster, in the middle of the plot, represents customers with average incomes and average spending scores. The other clusters represent customers with low incomes and low spending scores, low incomes and high spending scores, high incomes and low spending scores, and high incomes and high spending scores, respectively. The cause of the clusters is likely due to the fact that customers with higher incomes tend to spend more money than customers with lower incomes. However, there are some outliers in the data, such as the cluster of customers with low incomes and high spending scores, or the cluster of customers with high incomes and low spending scores. These outliers could be due to a variety of factors, such as customers with high incomes who are thrifty, or customers with low incomes who make impulsive purchases.*

Remove the word respectively from the caption.

*The clusters in this scatter plot represent different spending habits of customers based on their annual income. The largest cluster, in the middle of the plot, represents customers with average incomes and average spending scores. The other clusters represent customers with low incomes and low spending scores, low incomes and high spending scores, high incomes and low spending scores, and high incomes and high spending scores. The cause of the clusters is likely due to the fact that customers with higher incomes tend to spend more money than customers with lower incomes. However, there are some outliers in the data, such as the cluster of customers with low incomes and high spending scores, or the cluster of customers with high incomes and low spending scores. These outliers could be due to a variety of factors, such as customers with high incomes who are thrifty, or customers with low incomes who make impulsive purchases.*

For observations in the wider scope, we see that Tier 1 leans more to a style of a witty one-liner that makes no references to the analysis measurements provided. Tier 2 and Tier 3 captions start to include more inquiry about the data set's domain. The new caption generations for Tier 3 are mainly answers appended to the previous caption, unless it was explicitly instructed to modify something else.

## 5 USER STUDY

We conducted a user study to empirically test how engaging the captions were and to see which level of human effort would engaging the most engaging ones. A Qualtrics survey was sent out by email and Slack requesting for anonymous responses. Due to a tight time schedule, participants were limited to a scientific computing background (as this project is supported by an REU site) and 11 responses were used after a two-day wait period. Participants were shown four scatter plot visualizations in turn and for each they were asked to rank three captions (one from each tier) based on given sentiment/quality, choose the most engaging caption and optionally provide an explanation. The sentiments/qualities were relevance (does the caption match the visualization and analysis shown), repetitiveness (is the caption redundant and/or obvious in its information) and novelty (does the caption mention new information that is not shown in the visualization). Users were also given an option of None and instructed to place this option on top if they did not feel that any of the captions were of that sentiment/quality. The scatter plot visualizations contained two linear regressions and two cluster visualizations. The visualizations in each category were diverse in their metadata (e.g. one linear regression visualization had one outlier, the other had no outliers). Visualizations and captions were not randomized; captions were shown all at once in order of Tier 1 to Tier 3.

### 5.1 User Study Preliminary Results

The user study had 11 participants, shown 4 visualizations and 3 captions for each one. Most likely due to unclear instructions, some participants ended up placing the None option in the middle of their rankings. When aggregating the responses, we assume that this placement meant that they felt the captions below were not of the sentiment/quality. For example, if a participant had a most-to-least ranking of Caption 1-None-Caption 2-Caption 3, we added a count of Caption 1 for most relevant and did not include the rest in our results. There were also several responses that voted none of the captions for a sentiment/quality. Therefore, not all of the counts for each tier will add up to exactly 44 votes.

Although our preliminary results only consist of 11 responses, there are visibly clear trends of the non-none votes. Tier 2 and Tier 3 prompts yielded overwhelmingly more engaging captions than the Tier 1 prompts. Thus prompts with any amount of human effort were more engaging than no prompts with no human effort at all. The top left chart in Figure 5.1 shows the distribution of votes for each tier after aggregating the rank votes for all visualizations.

Results for the individual sentiment/qualities rankings also reveal a trend with level of human effort. Tier 1 with no human effort is ranked least relevant as shown in the top right chart with Tier 2 and Tier 3 are close in relevancy. The bottom charts show that as the level of human effort increases, the more repetitive but more novel the caption becomes. This could be a possible trade-off point between engagement and non-engagement. As with the engagement poll, votes were aggregated from all the visualizations.

## 6 DISCUSSION

We provide our explanation for these trends here. The wide gap between Tier 1 and Tier 2-3 in ranking for the most engagement could be linked to participants' preference for more elaborate and in-depth captions. Both Tier 2 and Tier 3 provided explanation for key parts in the visualizations. The novel and repetitive ranking trend also seems to have a logical reason in that Tier 3 contains the most information due to its answering of user requests, but in the process might add something undesired or already present in the visualization.

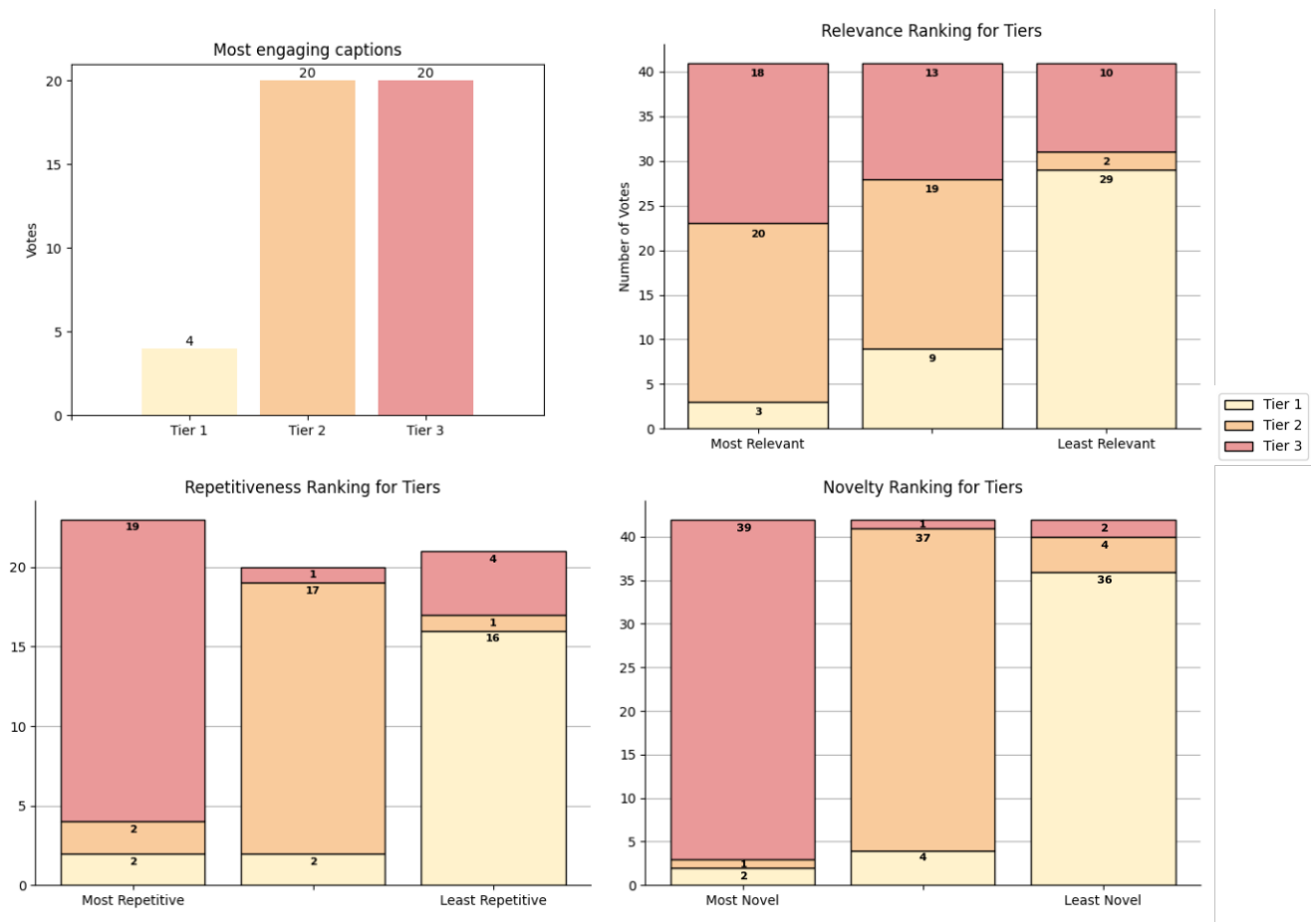

Figure 2: Aggregated results as stacked bar charts

What is most surprising to us is the ranking of Tier 1 as the least relevant captions. We suspected that Tier 3 would take this place due to its tendency towards repetitiveness and susceptibility to GPT-3's truthfulness hole mentioned in Section 4. A possible reason for this unexpected result might again be its lack of detail compared to Tier 2 and Tier 3.

## 7 FURTHER WORK

Visualizations are found in many works, from scientific papers to journalistic articles, and consequently their captions match a specific style of prose and content depth. Such specifications might require different prompts than what we have experimented with. Changes to the prompt generation process can be as minor as a synonym or completely different wording. Adjustment of GPT-3's hyperparameters to aid caption generation can also be studied to see how it affects the engagement levels.

Given this is our first experiment, a more concrete study into what makes a caption engaging would provide direction into what prompt engineering techniques besides human effort can be used. Having a definition of what makes something engaging would allow for better questions relating to engagement in the user study. Other parts of the user study could be enhanced, such as clearer instructions with ranking, randomization of captions, and excluding incoherent visualizations-captions as mentioned in Section 4. A more refined experimentation process, such as employing experts to interact with GPT-3 especially in Tier 3, could better serve as examples to use in the user study. Analysis of the user study can include more narrow results for each specific visualization that was shown rather than aggregation.

As mentioned in Section 6, GPT-3 is incapable of fact-checking itself. Further work can be done in figuring what prompt engineering techniques can be used to fill this truthfulness hole in GPT-3 for caption generation. We only look at human effort as our form of prompt engineering to make engaging captions, but there can very well be other prompt engineering techniques out there.

### ACKNOWLEDGMENTS

This research was supported by the Data + Computing = Discovery! REU site, which is sponsored by the NSF under grant 1950052.